# Applying Ensemble Models based on Graph Neural Network and Reinforcement Learning for Wind Power Forecasting

Hongjin Song[a], Qianrun Chen[b], Tianqi Jiang[c], Yongfeng Li[d], Xusheng Li[e], Wenjun Xi[f]
and **Songtao Huang**[b,*]
[a]Northwestern University
[b]Lanzhou University
[c]South China University of Technology
[d]Xi'an Jiaotong University
[e]Cloud Gansu Technology Co., Ltd
[f]Deer AI (Gansu) Technology Co., Ltd
hongjinsong2024@u.northwestern.edu, {chenqr2024, huangst21}@lzu.edu.cn,
cttqjiang@mail.scut.edu.cn, yflizdj@stu.xjtu.edu.cn, lixs666@gmail.com,
wenjunxi1997@gmail.com

## Abstract

Accurately predicting the wind power output of a wind farm across various time scales utilizing Wind Power Forecasting (WPF) is a critical issue in wind power trading and utilization. The WPF problem remains unresolved due to numerous influencing variables, such as wind speed, temperature, latitude, and longitude. Furthermore, achieving high prediction accuracy is crucial for maintaining electric grid stability and ensuring supply security. In this paper, we model all wind turbines within a wind farm as graph nodes in a graph built by their geographical locations. Accordingly, we propose an ensemble model based on graph neural networks and reinforcement learning (EMGRL) for WPF. Our approach includes: (1) applying graph neural networks to capture the time-series data from neighboring wind farms relevant to the target wind farm; (2) establishing a general state embedding that integrates the target wind farm's data with the historical performance of base models on the target wind farm; (3) ensembling and leveraging the advantages of all base models through an actor-critic reinforcement learning framework for WPF.

## 1 Introduction

With the development of the economy and science, human demand for energy is increasing daily. Over the past few decades, humans have heavily relied on fossil fuels for energy. However, the extensive use of fossil fuels has significantly harmed nature and humanity, contributing to the greenhouse effect and global warming. Currently, almost 99% of the world's population breathes air of a quality below World Health Organization standards. Approximately 8 million people die annually from household (indoor) and ambient (outdoor) air pollution. The World Health Organization estimates that in the Western Pacific region alone, 2.2 million individuals die each year due to air pollution (WHO, 2022). Consequently, as a clean, cheap, and renewable energy compared to traditional fossil fuels, wind power has attracted wide attention from governments and research institutions. According to the Global Wind Report 2022, the year 2021 is one of the best on record for the wind energy industry, with 93.6 GW of new capacity installed globally. At present, the total global wind power capacity stands at 837GW. The use of wind energy can help people avoid emitting carbon dioxide equivalent to South America's annual carbon emissions, about 1.2 billion tons (Council, 2022).

As a significant component of smart grids, wind power plays a crucial role in the generation and supply of energy (Wang et al., 2021). Wind power forecasting (WPF) is essential for the rational scheduling and optimization of wind power in one or more wind farms. Due to the significant instability and unpredictability of wind power supply, it has always been a complex challenge for researchers to achieve high-precision wind prediction.

Based on the time scale of prediction, WPF can be broadly categorized into four types (Chang et al., 2014; Wang et al., 2011):

- Very-short-term forecasting: The time range is a few minutes to an hour.
- Short-term forecasting: The time range is a few hours to a day.
- Standard-term forecasting: The time range is a few days to a week.
- Long-term forecasting: The time range is a few weeks or more ahead.

Very-short-term forecasting is mainly used for the regular cleaning of the electricity market and real-time operation of the grid. Short-term forecasting plays a significant role in financial market scheduling and informed load decision-making. Standard-term forecasting is employed for energy valuation and grid system control, while long-term forecasting mainly guides optimal cost assessment and feasibility method design for the wind farm planning.

From a modeling perspective, wind power forecasting methods can be subdivided into traditional methods (Sideratos & Hatziargyriou, 2007; Shi et al., 2011) and machine learning methods (Demolli et al., 2019; Jørgensen & Shaker, 2020). Although current methods have achieved significant success for WPF, numerous variables (temperature, altitude, position, humidity, pressure, etc.) could influence the results.

Each methods has its advantages in specific aspects. Therefore, leveraging and ensembling these models to optimize prediction accuracy for WPF is a promising method.

In this paper, we model all wind turbines within a wind farm as graph nodes in a graph constructed based on their geographical locations. Consequently, we propose an ensemble model based on graph neural network and reinforcement learning (EMGRL) for WPF. Our approach includes: (1) applying graph neural networks to capture the time-series data from neighboring wind farms relevant to the target wind farm; (2) establishing a general state embedding that integrates the target wind farm's data with the historical performance of base models on the target wind farm; (3) ensembling and leveraging the advantages of all base models through an actor-critic reinforcement learning framework for WPF. We demonstrate that the EMGRL model outperforms state-of-the-art (SOTA) baselines by up to 12.89% in comparison trials on open datasets for WPF.

The remainder of this paper is organized as follows: Section 2 introduces related works on WPF, including traditional and machine learning methods. Section 3 summarizes the basic technologies and depicts the problem definition of this paper. Section 4 proposes the ensemble model based on graph neural networks and reinforcement learning (EMGRL). Section 5 reports the results of comprehensive experiments on real-world wind farms datasets, comparing EMGRL with baseline methods. Finally, conclusions are presented in Section 6.

## 2 Related Works

### 2.1 Traditional Methods

Compared to machine learning methods, traditional methods emphasize wind power prediction through rule-based modeling. These traditional approaches can be classified into persistence, physical, and statistical methods.

#### 2.1.1 Persistence Methods

The general idea of persistence forecasting is to use the average value of historical wind power to predict that future wind power will be the same as the current value. Persistence forecasting is typically used as a baseline measure of the accuracy of other prediction methods rather than as a practical method, especially for short-term forecasts (Bludszuweit et al., 2008).

$$P_{t+\delta t} = P_t \tag{1}$$

where the wind power $P_{t+\delta t}$ at a future time step $\delta t$ is predicted to be the same as the current wind power $P_t$.

Since the persistence method pays little attention to parameters, it is not reasonable to use it when the forecast duration is longer than a few hours.

#### 2.1.2 Physical method

The physical method typically uses historical weather data from the wind farm, such as temperature, surface pressure and the number of obstacles, etc., to forecast wind speed and create wind power maps using wind turbine power curves (Santhosh et al., 2020). The most widely used physical approach is Numerical Weather Forecasting (NWP). NWP predicts future weather phenomena at a specific time by solving mechanical equations that describe the evolution of weather under certain initial conditions. The major NWP models currently used in the industry include the GRAPES model (published by the China Meteorological Administration), the WRF model (published by the National Meteorological Research Department), and the IFS model (published by the European Meteorological Forecasting Agency) (Wu et al., 2022).

Pelikan et al. used mesoscale resolution NWP to study wind power prediction as early as 2010. After validating with actual electricity data in the Czech Republic, they found that the model could make accurate predictions within 72 hours (Pelikan et al., 2010). To address the low performance of the NWP prediction algorithm, Wang et al. (Wang et al., 2019) proposed an NWP speed correction algorithm based on the Sequence Transfer Correction Algorithm (STCA). This algorithm demonstrated excellent generalization ability for a variety of models, proving the physical method's applicability.

The results obtained by the physical method are consistent with physical laws and relatively accurate, which may be the best method for standard and long-term wind power prediction (Hanifi et al., 2020). However, implementing physical methods requires excessive computational time and system resources, making them unsuitable for short-term forecasting.

### 2.1.3 Statistical method

Statistical methods for wind power forecasting include the autoregressive integrated moving average model (ARIMA) (Nelson, 1998), Kalman filter (Welch, 2020) and Bayesian model (Pole et al., 2018), which aim to determine the relationship between input variables and output using historical datasets.

The ARIMA technique is commonly utilized in time series forecasting tasks, including wind power forecasting. The usual steps of ARIMA include (1) model identification, (2) model ordering, (3) parameter estimation, and (4) prediction (Kavasseri & Seetharaman, 2009).

Yatiyana et al. (Yatiyana et al., 2017) used the ARIMA model to establish a wind power forecasting model in Western Australia, significantly improving the reliability and quality of the wind power system. Although ARIMA accurately captures the conditional mean, it excludes the time-varying variations of wind power generation. Tian et al. (Tian et al., 2018) proposed a novel wind power forecasting algorithm based on the ARIMA-LGARCH model, considering the impact of random wind power volatility on prediction accuracy. To analyze the non-stationary and autocorrelation aspects of wind power time series data, the ARIMA and Logarithmic Generalized Autoregression Conditional Heteroscedasticity (LGARCH) models were utilized. The model's feasibility and efficacy were confirmed by comparing the predicted and actual wind power values in the wind farm. The prediction model's validity was enhanced to estimate and predict the parameters more precisely. Gupta et al. (Gupta et al., 2019) studied the advantages of the Generalized Autoregressive Conditional Heteroskedasticity (GARCH) model and proposed a hybrid ARIMA-GARCH model for wind power prediction. This model combines ARIMA-based conditional mean prediction with GARCH-based conditional variance prediction. Simulation results from three wind farms in Australia, evaluated using MAE and RMSE, show that the proposed model outperforms the classic ARIMA model.

The kalman filter is a state estimation algorithm that combines prior experience with measurement update, widely used in prediction tasks. Bossanyi et al. (Bossanyi, 1985) was the first to use the Kalman filter for wind forecasting. Salgado et al. (Salgado et al., 2018) devised a method to calculate the average hourly wind speed over a 24-hour period using a multimodal Kalman filter. A clustering algorithm was used to obtain the wind speed characteristic curve of the submodel. After verification, the proposed prediction model outperformed other statistical methods in terms of accuracy. The Kalman filter method uses the linear system state equation to process observed data and optimally estimate the system state. However, when there is nonlinear fluctuation in the data, the prediction effect of the model is not ideal.

Wang et al. (Wang et al., 2017) proposed a generalized multi-core regression model and introduced the variational Bayes method to improve the model. This was done to address the shortcomings of traditional wind power prediction models, which are not sensitive to high-resolution data and have assumptions inconsistent with error distribution. Research findings in China suggest that the proposed model can estimate probability more accurately and has high potential for adoption in modern power systems. Wang et al. (Wang et al., 2020a) developed a hybrid wind power prediction method based on Bayesian Model Averaging and Ensemble Learning (BMA-EL). Firstly, training sets with the same distribution are imported into three basic learners to train the model. The combined policy is then trained based on the validation set output. Finally, BMA-EL generates the wind power prediction result through this output. Compared to other algorithms, BMA-EL is shown to estimate wind power output with high accuracy and stability across various environmental situations.

Statistical methods are easier to model and less expensive to develop than physical methods. However, as the forecasting time increases, the error increases sharply, making most statistical methods suitable only for short-term predictions. Another disadvantage is that these statistical models do not account for the influence of time delays, making the integration of wind power time series into the forecasting model problematic (Deng et al., 2020).

## 2.2 Machine Learning methods

Machine learning methods can be further classified into traditional and deep learning approaches. Traditional machine learning methods focus on modeling and predicting wind power through feature engineering. In contrast, deep learning methods primarily use neural networks to directly output prediction results based on big data.

### 2.2.1 Traditional Machine Learning methods

The goal of machine learning is to create algorithms that learn from data and generate predictions and allow computers to update themselves. Because machine learning algorithms can accurately characterize the behavior of datasets, input features are modeled based on expected outputs, and output features are predicted based on previous data. Hence, machine learning is one of the effective methods for wind power prediction (Demolli et al., 2019).

The SVM method performs well in wind power prediction tasks. Zeng et al. (Zeng & Qiao, 2011) proposed a statistical method for wind power prediction based on SVM. The model predicts the wind speed first, and then the wind power is expected using the wind turbine's power and wind speed characteristics. According to simulation findings based on data from the National Renewable Energy Laboratory (NREL), this model has higher accuracy for very short-term and short-term WPF compared to models based on the continuous model and radial basis function neural network. Many scholars have proposed improved SVM methods for wind power forecasting. Liu et al. (Liu et al., 2020) proposed a support vector machine (JAYA-SVM) short-term wind speed prediction model based on the JAYA algorithm. The hyperparameters of the SVM are optimized using the JAYA optimization algorithm using the most representative features in the input data.

In addition to SVM, traditional machine learning algorithms such as KNN (Mahaseth et al., 2022), XGBoost (Jiading et al., 2022) and Random Forest (RF) (Lahouar & Slama, 2017; Shi et al., 2018) have also been used in wind power prediction tasks. However, traditional machine learning methods require manual extraction and cleaning of data, followed by feature engineering. The selection of features largely determines the model's effectiveness. Therefore, traditional machine learning methods have significant limitations in the application of flexible wind power prediction scenarios.

### 2.2.2 Deep Learning methods

Deep learning can refine data using representation and machine learning models without the need for preprocessing steps such as feature selection, dimension compression, and format conversion. The primary advantage of deep learning over traditional methods is its ability to automatically extract features. Advances in modern computing power and neural network theory have provided extensive application potential for deep learning models.

Convolutional neural networks (CNNs) can be combined with radial basis function neural networks (RBFNNs) for wind power generation prediction (Hong & Rioflorido, 2019). CNNs are used to extract wind power features through convolution, kernel, and pool operations, while supervised RBFNNs handle uncertain features. Testing has shown that this method consistently achieves excellent wind power forecasts, with values exceeding 0.9 in both winter and summer, indicating outstanding 24-hour-ahead predictions. Devi et al. (Devi et al. 2020) introduced an enhanced long short-term memory network, the Augmented Forgetting Gate Network (LSTM-EFG), to forecast subsequence data extracted using Ensemble Empirical Pattern Decomposition (EEMD). The cuckoo search optimization algorithm (CSO) was incorporated into the parameter design process. Experimental results demonstrate that this model significantly improves prediction accuracy and avoids the defects of traditional models. Kisvari et al. (Kisvari et al., 2021) proposed a method based on a gated recursive deep learning model combined with traditional machine learning hyperparameter tuning. They successfully developed a novel gated recurrent unit (GRU) neural network, which includes 12 features such as generator temperature, gearbox temperature, and wind speed at different heights. In practical tests, the GRU's prediction accuracy surpassed that of LSTM and was less sensitive to noise in SCADA datasets. The attention mechanism also proves valuable in wind power prediction. Niu et al. (Niu et al. 2021) introduced an attention-based gated recurrent unit sequence model (AGRU) to improve prediction performance, using hidden GRU modules to connect and activate tasks across different prediction steps.

A new data-driven model based on the principles of deep learning-convolutional long short-term memory (CLSTM), evolutionary algorithms, a neural architecture search process, and integrated deep reinforcement learning (RL) strategy was proposed by Jalali et al. (Jalali et al., 2021). Initially, mutual information is used to extract features from unprocessed wind power time series data. The architecture of the deep CLSTM model is then optimized through the neural architecture search procedure. Finally, the RL algorithm-based AI model is integrated to minimize the wind power prediction gap between two datasets. This algorithm demonstrates excellent seasonal performance when compared to 14 state-of-the-art models. Chen et al. (Chen et al., 2021) proposed a dynamic integrated wind speed forecasting model based on deep reinforcement learning, considering the time-varying characteristics of wind speed series. The first part involves using real-time wavelet packet decomposition enhanced deep echo state network to construct the basic model with different vanishing moments. The second part determines the weight allocation of the basic learner using a multi-modal optimization method. The third part embeds the non-dominated solution of combined weights in a deep reinforcement learning environment, enabling dynamic selection. By carefully designing the reinforcement learning environment, a dynamic non-dominated solution can be selected in each prediction round according to the time-varying properties of wind speed.

Additionally, combining various models to improve prediction performance through ensemble methods integrates the advantages of different models and offers good generalization ability (Du et al., 2019; Dong et al., 2021; Jaseena & Kovoor, 2021; Duan et al., 2022).

## 3 Problem Preliminary

### 3.1 Ensemble learning

Ensemble learning first extracts the characteristics of the data through a series of prescribed mathematical methods. Based on the obtained data characteristics, various algorithms are selected to generate weak prediction results. Finally, ensemble learning acquires new knowledge and achieves better prediction results through an adaptive voting scheme (Dong et al., 2022).

The comprehensive structure of ensemble learning is illustrated in Figure 1. Ensemble learning methods can generally be subdivided into Bootstrap aggregating (Bagging), Boosting, and Stacking. The bagging method randomly samples training subsets from the training set and then uses these subsets to perform parallel training while integrating the base learners into the ensemble model. The Boosting method first trains a basic learner with better performance from the initial training set, then adjusts the sample distribution according to the performance of the basic learner to increase the weight of the samples incorrectly predicted by the basic learner in the previous training. This process is repeated for the next basic learner with the optimized sample distribution, and the operations continue until completion. Finally, a weighted combination of the basic learners is performed. The Stacking model builds a meta-model using predictions from multiple base models to generate the final forecast. The Stacking model is composed of multiple layers, with each layer consisting of several machine learning models. The predictions from these models are used to train the next layer model.

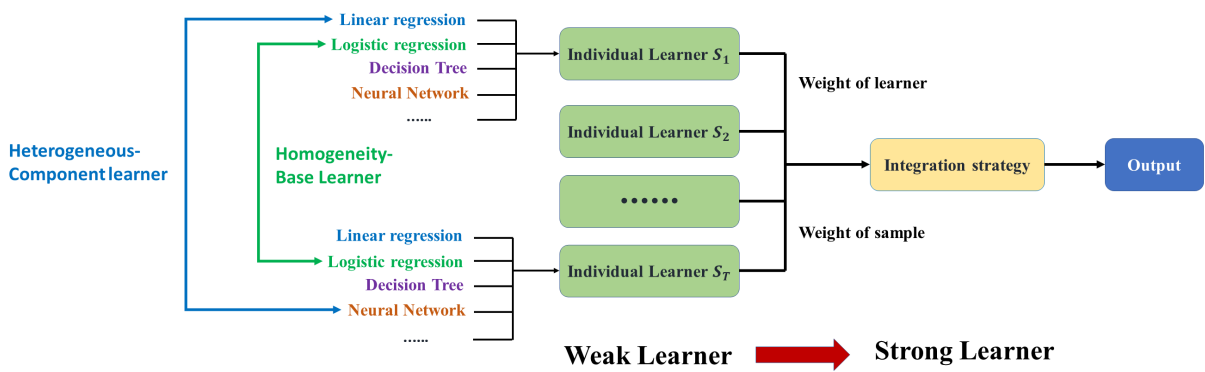

Figure 1: The overall framework of ensemble learning.

## 3.2 Graph neural network

Graph Neural Network (GNN) is an emerging framework that uses deep learning to directly learn from graph-structured data. Its excellent performance has attracted significant attention and in-depth exploration from scholars. By formulating specific strategies for the nodes and edges of the graph, GNN transforms the graph data into a standardized representation, which is then input into neural networks for training. This approach has achieved remarkable results in information aggregation, action recognition, and time series prediction.

GNNs are typically built on the message-passing neural network framework. As shown in Figure 2, the GNN message-passing process can be divided into three stages:

(1) Message Computation

In the Message Computation stage, function $MC^l$ is used to calculate the information value $m_u^l$ of node $u$ in layer $l$

$$m_u^{\{l\}} = MC^l(h_u^l) \qquad (2)$$

(2) Message Aggregation

In the Message Aggregation stage, the function $MA^l$ is used to integrate the information of node $u_i$ in the neighborhood of node $v$ located in layer $l$

$$a_v^{\{l\}} = MA^l(\{m_{u_i}^l, u_i \in N(v)\}) \qquad (3)$$

(3) Message Update

In the Message Update stage, the function $MU^l$ updates node $v$ at layer $l$ based on the neighborhood aggregation information $a_v^l$ and its own information $m_v^l$

$$h_v^{\{l+1\}} = MU^l(a_v^l, m_v^l) \qquad (4)$$

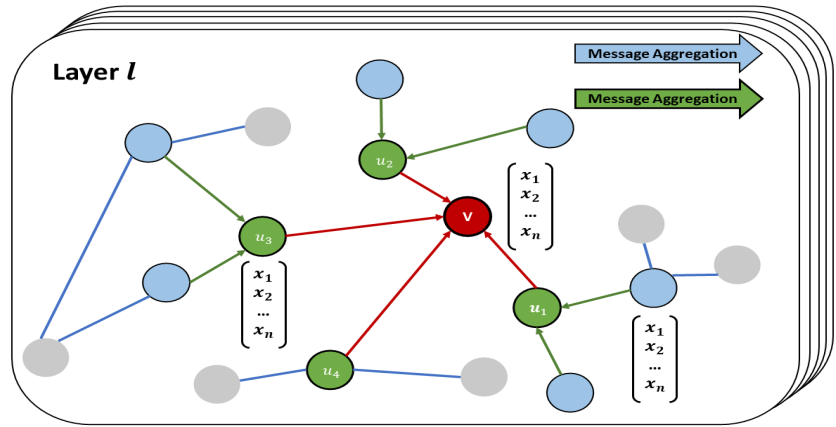

Figure 2: The Message aggregation process in GNN for node $v$.

## 3.3 Reinforcement learning

Reinforcement learning is a field within machine learning (Sutton et al., 1998). It encompasses a class of learning methods that learn by interacting with the environment and using feedback from it. With the recent development of artificial intelligence (AI), reinforcement learning has emerged as one of the three main machine learning paradigms, alongside supervised learning and unsupervised learning, due to its powerful autonomous learning capabilities.

The core of reinforcement learning is to study the interaction between the agent and the environment. By continuously learning the optimal policy, the agent makes sequential decisions based on the current state of the environment to maximize cumulative reward. This process can be described as a Markov Decision Process (MDP), as shown in Figure 3.

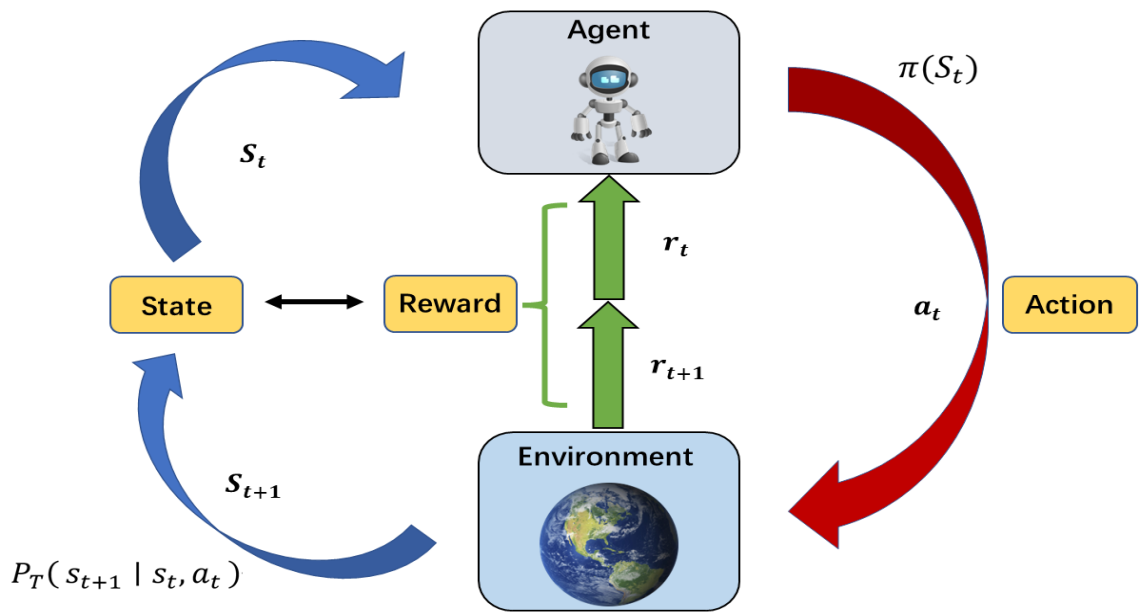

Figure 3: Markov Decision Process (MDP)

The parameter space of an MDP can be represented as a tuple $(S, A, P, R, \gamma)$, where: $S$ is a finite set of states with $s_t \in S$, where $s_t$ is the state at time $t$; $A$ is a finite set of actions, with $a_t \in A$, where $a_t$ is a specific action; $P$ represents the state transition probabilities, $P_T(s_{t+1} \mid s_t, a_t)$, predicting the next state $s_{t+1}$ based on the current state $s_t$ and action $a_t$. $P_T$ indicates the possibility that action $a_t$ will transition the state from $s_t$ to $s_{t+1}$; $R$ is the reward fuction, with $r_t \in R$, where $r_t$ is the immediate reward received after taking an action; $\gamma$ is the discount factor. The policy $\pi(s_t)$ represents the possibility of taking action $a_t$ based on the current state $s_t$. The agent's value function, $V(s_t)$, and action value function, $Q(s_t, a_t)$, evaluate the expected long-term reward the agent can receive in the state $s_t$.

## 3.4 Problem definition

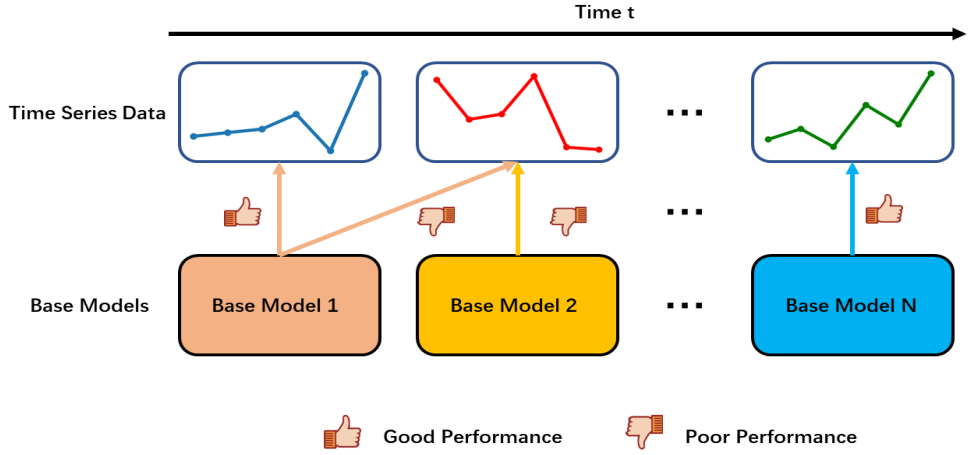

Figure 4: Ensemble learning model decision process.

The main objective of wind power prediction is to forecast wind power over a period based on historical wind power data from a specific area (wind farm) and the wind power characteristics of neighboring areas. For time series data, as shown in Figure 4, different base models often perform differently at the same stage of the data. Therefore, identifying the model that performs best at a certain stage is more valuable and easier to implement than directly using a single learner for predictions. This strategy allows us to find the best model to adapt to changing data distributions and make accurate predictions.

We aim to solve two problems: (1) What is the next value in the time series data? (2) Which base model is better at predicting the next value? For problem (1), we use various time series prediction methods as base models to make simultaneous predictions. For problem (2), we use reinforcement learning to assign weights to each base model, reflecting their respective contributions to the prediction results in ensemble learning. In this prediction task, it is necessary to splice and embed the data of each wind farm and the loss data of each learner, then calculate the weight of each base learner using reinforcement learning, and finally make the prediction.

$$W_{\text{final}} = Predict(b_1, b_2, b_3 \cdots b_n), n \in N \tag{5}$$

where $Predict$ stands for the prediction function, $b_i$ stands for base model $i$ in an ensemble model, and $N$ represents the number of base models selected.

## 4 Methods

In this section, we introduce and detail an ensemble model for wind power forecasting (WPF) based on a graph neural network and reinforcement learning (EMGRL).

## 4.1 The overview of EMGRL

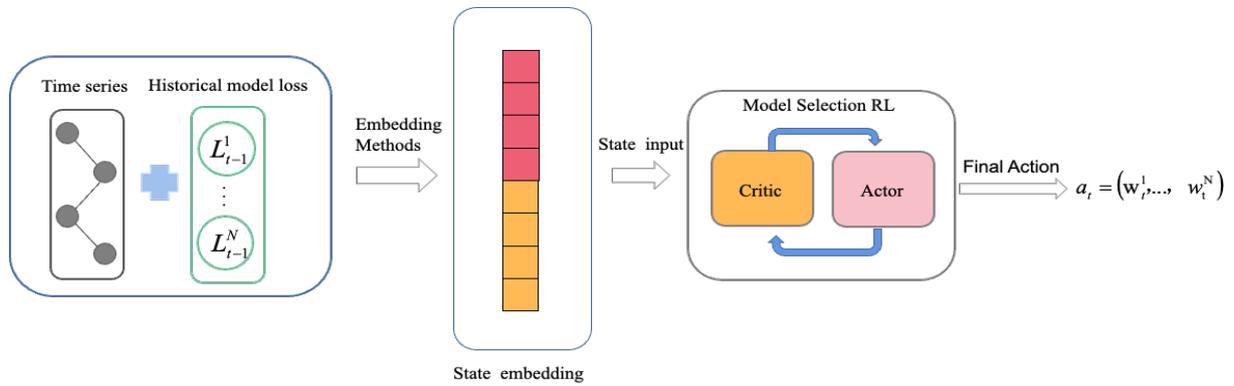

Figure 5: The overall framework of EMGRL model.

The overall framework of the EMGRL model is shown in Figure 5. Firstly, we propose a method for generating state embeddings that integrates temporal data and spatial features of wind farms using a dilated convolutional neural network and a graph neural network, further combined with model loss to generate embedding vectors. Secondly, we input the obtained embedding vector into an actor-critic reinforcement learning model, which outputs the weight of each base model in the ensemble model through iterative training to achieve optimal results.

## 4.2 State embedding generation

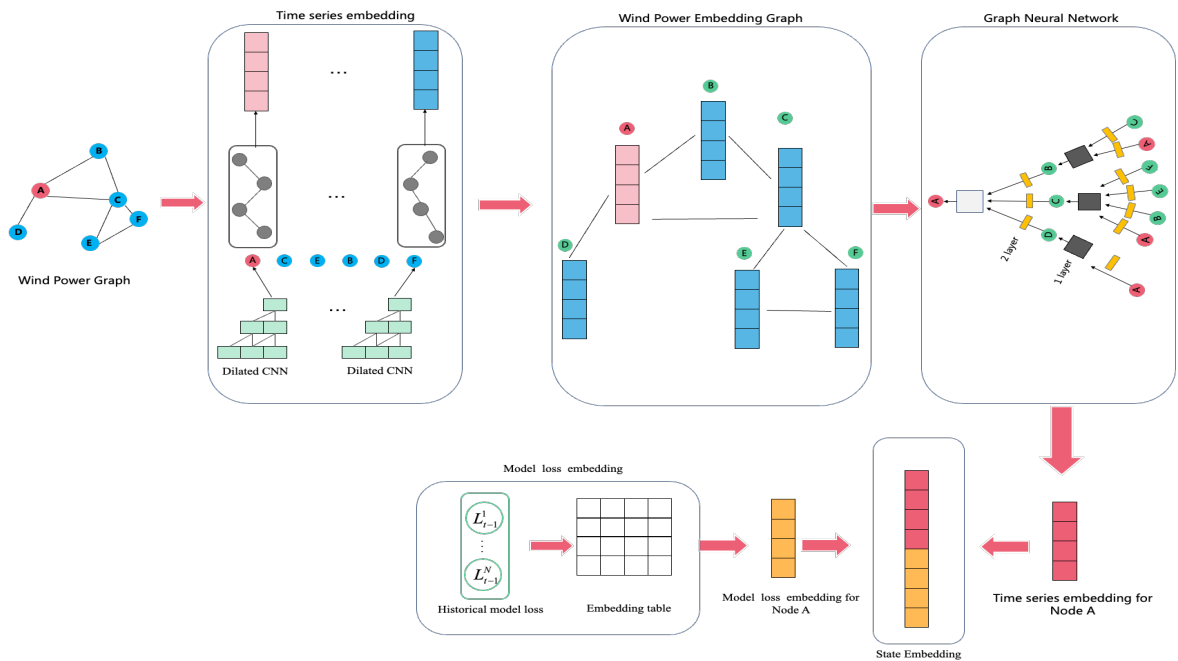

Figure 6: A flowchart for generating state embedding vectors.

The state embedding consists of two parts: spatio-temporal sequence embedding and model loss embedding. For the spatio-temporal sequence embedding part, we obtained each wind farm's compressed time sequence data vector through a dilated convolutional neural network (as shown in Figure 6). Based on time series data, we further consider the spatial characteristics of wind power. We input the timing vectors of all wind farms, including geographical location information, into the GNN network. As shown in Figure 6, we compute the embedding of wind farm A among wind farms (A, B, C, D, E, F). Through training, the GNN network can simultaneously integrate the information of neighboring wind farms into wind farm A's time sequence data vector, obtaining the spatio-temporal sequence embedding of wind farm A. The computation of the spatio-temporal sequence embedding for other wind farms follows the same process as for wind farm A. We denote the space-time embedding vector of wind farm A as $STSE_A$, which represents the raw input data of the wind farm. The following expression shows the relationship between the two:

$$STSE_A = GNN(\{\text{Dilated} - CNN(t_k)\}), k \in \{A, B, C, D, E, F\} \tag{6}$$

For the mode loss embedding part, we integrate all the basic models in the learning model, conducted training on the historical data of all wind farms, and learned the internal parameters of each basic model. Using wind farm A as an example, we obtained the predicted losses of all basic models on wind farm A in each training session, spliced the losses, and compressed them through a dilated convolutional neural network to obtain the model loss embedding of wind farm A. Let $L_{t-1}^i$ represent the loss of basic model $i$ on the time sequence data of wind farm A at time $t-1$, and let $N$ represent the number of basic models. The computation of this method is shown in the formula:

$$MLE_{t-1}^A = MLP(\{L_{t-1}^i\}, i \in \{1, \cdots, N\}) \tag{7}$$

Hence, we obtain the embedding vector $SE_A$ of wind farm A through splicing:

$$SE_A = (STSE_A, MLE_{t-1}^A) \tag{8}$$

## 4.3 Training and Prediction Pipeline

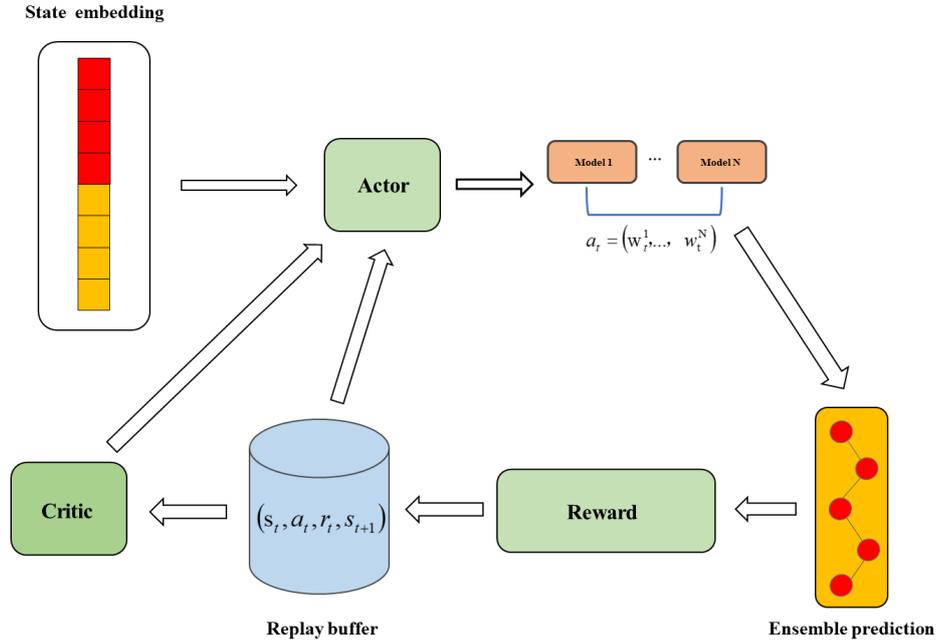

Figure 7: Reinforcement learning model training process

We outline the following processes for the training and prediction phases of the reinforcement learning model (as shown in Figure 7). Our reinforcement learning algorithm is based on the Actor-critic framework. At timestep $t$, the state $s_t$ describes the information of the input time series and the historical performance of the base models, represented as $SE_k, k \in \{A, B, C, D, E, F\}$. We input $SE_k$ into Actor module, which selects and outputs the appropriate actions according to a predetermined policy. The output value of the actor is $a_t = (W_t^1, \ldots, W_t^N)$, and each element of the vector corresponds to the weight of the $N$ base models in the wind power prediction results. Based on the weight $a_t$, we can output the ensemble prediction $y_t$ of wind farm k through the ensemble model. The reward $r_t$ is calculated by comparing $y_t$ with the real data. Meanwhile, we also obtain the next state $s_t'$ after the EMGRL model takes action $a_t$. We then save the transition $(s_t, a_t, r_t, s_t')$ into the replay buffer for later training. We update the actor $A(s_t)$ and critic $Q(s_t, a_t)$ using the sampled transitions from the replay buffer. Through continuous iterative training, highly accurate wind power prediction results for each wind farm can be achieved.

## 4.4 Algorithm

The EMGRL algorithm is:

---

**Algorithm 1** The EMGRL Algorithm

---

**Input** $SE_A$ as state $s$ for wind farm A:

**Initialize** θ for actor, ω for critic

   1:  **for** t=1 to T **do**:

   2:      Sample batch transitions $(s_t, a_t, r_t, s'_t)$ from replay buffer

   3:      Sample action $a'_t = \pi_\theta(s, a)$

   4:      Update critic:

$$w = \arg\min E_{\pi_\theta}\left(Q^{\pi(\theta)}(s_t, a_t) - Q_\omega(s_t, a_t)\right)^2$$

   5:      Update actor:

$$\theta = \theta + \alpha \log \pi_\theta(s_t, a_t) Q_\omega(s_t, a_t)$$

   7:      $a_t = a'_t; s_t = s'_t$

   8:  **end for**

---

## 5 Experiment

In this section, we conduct numerical experiments to demonstrate the performance of the EMGRL in wind power forecasting (WPF).

### 5.1 Dataset

### 5.1.1 NREL Dataset

In this experiment, we use an open-source dataset from the National Renewable Energy Laboratory (NREL) (Draxl et al., 2015). The NREL dataset contains wind data from 2010 to 2011 for four offshore wind farms (A, B, C, and D) located on the east coast of the United States. The data is sampled at 10-minutes intervals, with the wind turbine output power ranging from 0 to 16MW. We use the wind power data from 2010 for model training and the wind power data from 2011 for testing. The specific locations of the wind farms are shown in Fig. 8.

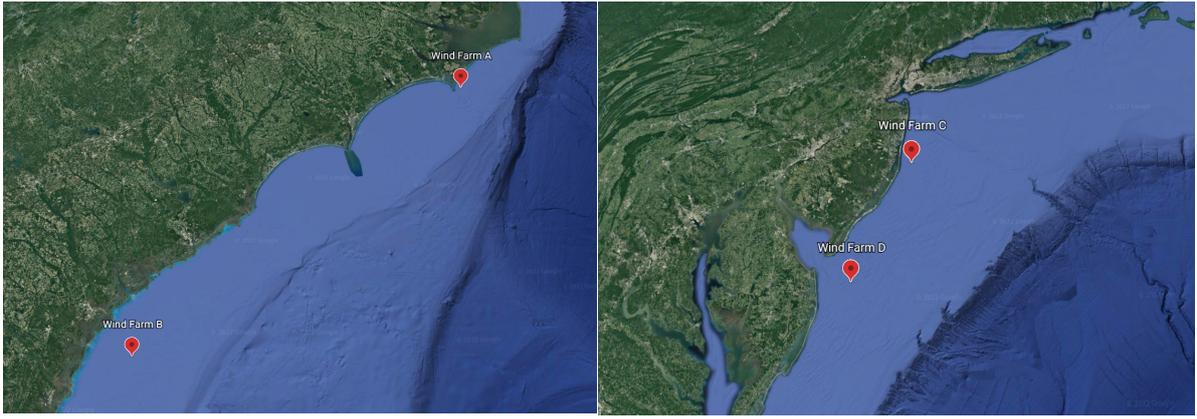

Figure 8: Abstract map of wind farm location in NREL Dataset.

### 5.1.2 GEFC Dataset

The second Dataset we use is from Global Energy Forecasting Competition (GEFC) 2012 - Wind Forecasting. The GEFC dataset includes three years of historical data from seven wind farms, spanning from the first hour of July 1, 2009 to the 12th hour of June 28, 2012. The historical data consists of hourly wind power measurements for each wind farm and forecasts of zonal and meridional wind components, wind speed, and direction for the next 1 to 48 hours. In this dataset, the period from July 1, 2009 to December 31, 2010, is used for model training, and the period from January 1, 2011 to June 28, 2012 is used for model testing. The wind power values are normalized to the interval between 0 and 1 to ensure that the original data features of the wind farm cannot be identified (Li & Chiang, 2016).

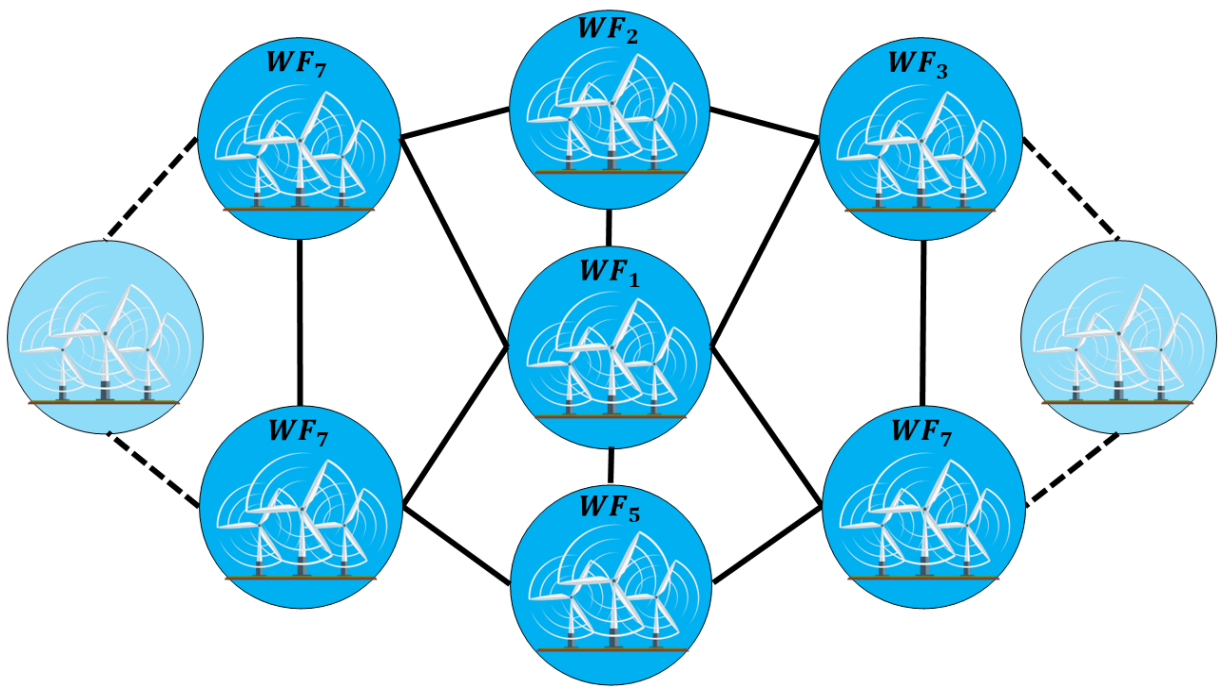

Figure 9: Abstract map of wind farm location in GEFC Dataset (Taking Wind Farm 1 as an example).

## 5.2 Baseline methods

- Autoregressive Integrated Moving Average (ARIMA) (Wang & Hu, 2015)
- Support Vector Regression (SVR) (Dhiman et al., 2019)
- LightGBM (Ren et al., 2022)
- Long short-term memory network (LSTM) (Ko et al., 2020)
- Superposition Graph Neural Network (SGNN) (Yu et al., 2020)
- COBRT (Guo & Chiang, 2016)
- SDAE (Yan et al., 2018)

## 5.3 Settings

In this paper, our integration model contains four basic models: ARIMA, LightGBM, LSTM, and SGNN. ARIMA is adept at thoroughly learning the linear characteristics of historical data, while LightGBM excels at handling the nonlinear aspects of time series data. LSTM can retain long-term data elements and learn data patterns, whereas SGNN analyzes data from a graph perspective to extract spatial features more effectively. Each of these four basic learning models brings unique strengths to wind power prediction. By leveraging their respective advantages and allocating weights reasonably, our integrated learning model aims to achieve the best comprehensive prediction results.

## 5.4 Evaluation

In this paper, we use Mean Absolute Error (MAE) and Root Mean Square Error (RMSE) as the evaluation metrics for the proposed model.

- MAE: Mean Absolute Error calculates the average of the absolute error between the predicted value $\hat{y}_i$ and the true value $y_i$, and is less sensitive to outliers.

$$MAE = \frac{1}{n}\sum_{i=1}^{n}|y_i - \hat{y}_i| \qquad (9)$$

- RMSE: Root Mean Square Error is the square root of the ratio of the square of the deviation between the real value $y_i$ and the predicted value $\hat{y}_i$ and the number of observations $n$, which can calculate the deviation between the predicted value and the real value. RMSE is sensitive to outliers in the data.

$$RMSE = \sqrt{\frac{1}{n}\sum_{i=1}^{n}(y_i - \hat{y}_i)^2} \tag{10}$$

**5.5 Results**

We compare the EMGRL model with seven baseline models (ARIMA, SVR, LightGBM, LSTM, SGNN, SDAE, and COBRT) under both datasets. For the NREL dataset, we compute the RMSE of the model on the data from four wind farms (A, B, C, and D) through multiple average calculations. For the GEFC dataset, we compute the MAE and RMSE of the model on the data from seven wind farms through multiple average results.

- Algorithm performance in NREL Datasets

As shown in Table 1, the EMGRL model achieves the lowest RMSE values for the four wind farms in the NREL dataset. Compared to SGNN, which performs the best among the basic models, EMGRL outperforms SGNN by approximately 9.62%, 10.44%, 6.57%, and 6.34% for wind farms A, B, C, and D, respectively. We visualize the RMSE values of EMGRL and each base model in Figure 10 and compared RMSE values of EMGRL and SGNN in Figure 11.

Table 1: MAE values of different baseline methods and EMGRL models on NREL Datasets.

| Model | Wind Farm A | Wind Farm B | Wind Farm C | Wind Farm D |
|---|---|---|---|---|
| ARIMA | 1.5465 | 1.4618 | 1.4594 | 1.3982 |
| SVR | 1.2954 | 1.2235 | 1.2300 | 1.1832 |
| LightGBM | 1.1698 | 1.1214 | 1.1283 | 1.0847 |
| LSTM | 1.3846 | 1.2982 | 1.3034 | 1.2750 |
| SGNN | 1.1520 | 1.1416 | 1.1184 | 1.0722 |
| SDAE | 1.3326 | 1.2143 | 1.1905 | 1.1761 |
| COBRT | 1.2517 | 1.1949 | 1.2109 | 1.1074 |
| **EMGRL** | **1.0509** | **1.0337** | **1.0494** | **1.0082** |

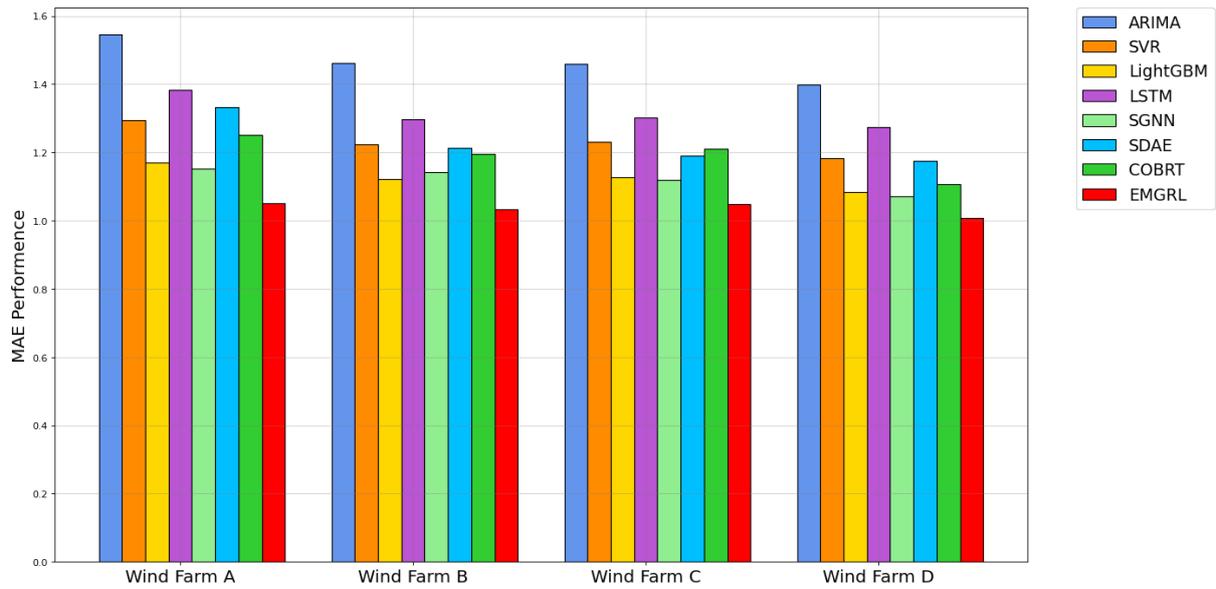

Figure 10: MAE performance comparison between EMGRL and multiple baseline methods on NREL Datasets.

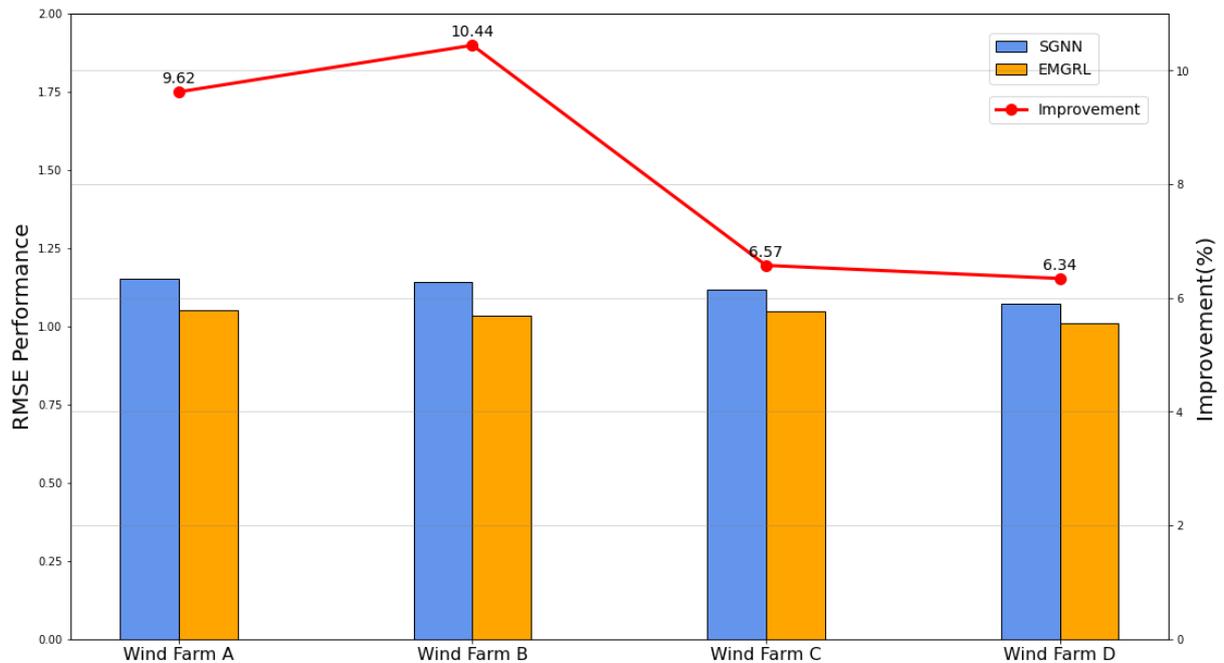

Figure 11: Improvement of EMGRL model on RMSE compared to the best-performing benchmark model SGNN in NREL Dataset.

- Algorithm performance in GEFC Datasets

As shown in Table 2, the EMGRL model also significantly outperforms other models in terms of MAE value, with the MAE for wind Farm 4 reaching a low of 0.1423. The MAE value of the EMGRL model is approximately 7.74% better than that of the best-performing model, SDAE, across the seven wind farms. We visualize and compare the MAE values for each model in Figures 12 and 13. Similarly, on the GEFC dataset (Table 3, Figure 14 and 15), the RMSE of the EMGRL model also achieves optimal results. These findings again demonstrate the superiority of our model, which performs well across multiple datasets.

Table 2: MAE values of different baseline methods and EMGRL models on GEFC Datasets.

| Model | Wind Farm 1 | Wind Farm 2 | Wind Farm 3 | Wind Farm 4 | Wind Farm 5 | Wind Farm 6 | Wind Farm 7 |
|---|---|---|---|---|---|---|---|
| ARIMA | 0.2215 | 0.2087 | 0.2150 | 0.2073 | 0.1969 | 0.1986 | 0.1976 |
| SVR | 0.1887 | 0.1772 | 0.1891 | 0.1760 | 0.1712 | 0.1682 | 0.1593 |
| LightGBM | 0.1666 | 0.1702 | 0.1630 | 0.1661 | 0.1637 | 0.1550 | 0.1469 |
| LSTM | 0.1754 | 0.1668 | 0.1768 | 0.1637 | 0.1523 | 0.1545 | 0.1564 |
| SGNN | 0.1792 | 0.1622 | 0.1795 | 0.1761 | 0.1655 | 0.1617 | 0.1502 |
| SDAE | 0.1650 | 0.1582 | 0.1663 | 0.1496 | 0.1717 | 0.1501 | 0.1522 |
| COBRT | 0.1734 | 0.1821 | 0.2073 | 0.1784 | 0.2029 | 0.1685 | 0.1794 |
| EMGRL | 0.1505 | 0.1492 | 0.1605 | 0.1423 | 0.1628 | 0.1435 | 0.1473 |

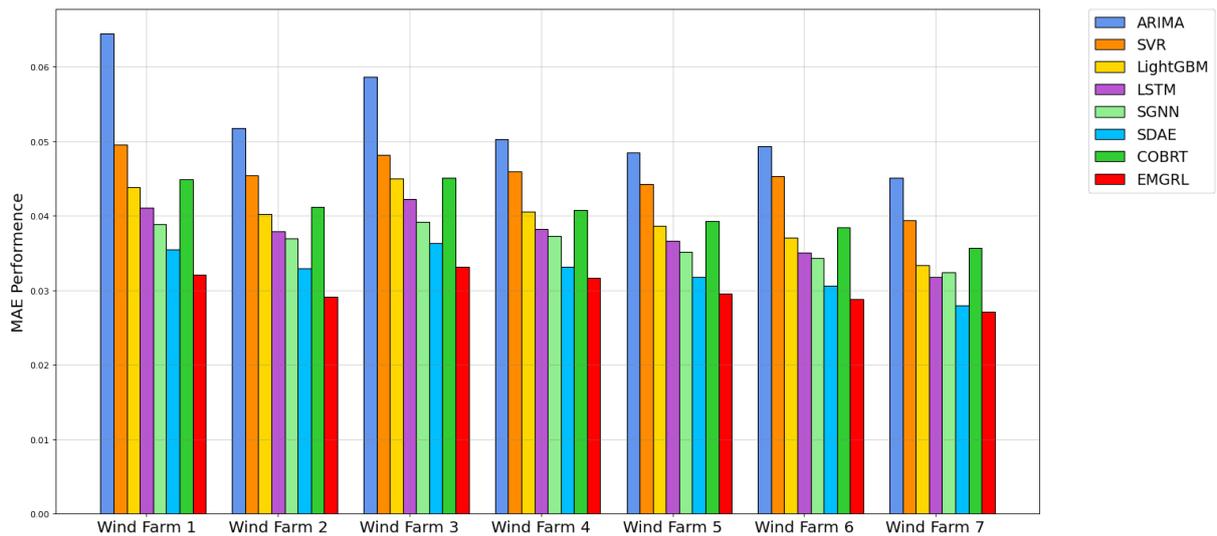

Figure 12: MAE performance comparison between EMGRL and multiple baseline methods on GEFC Datasets.

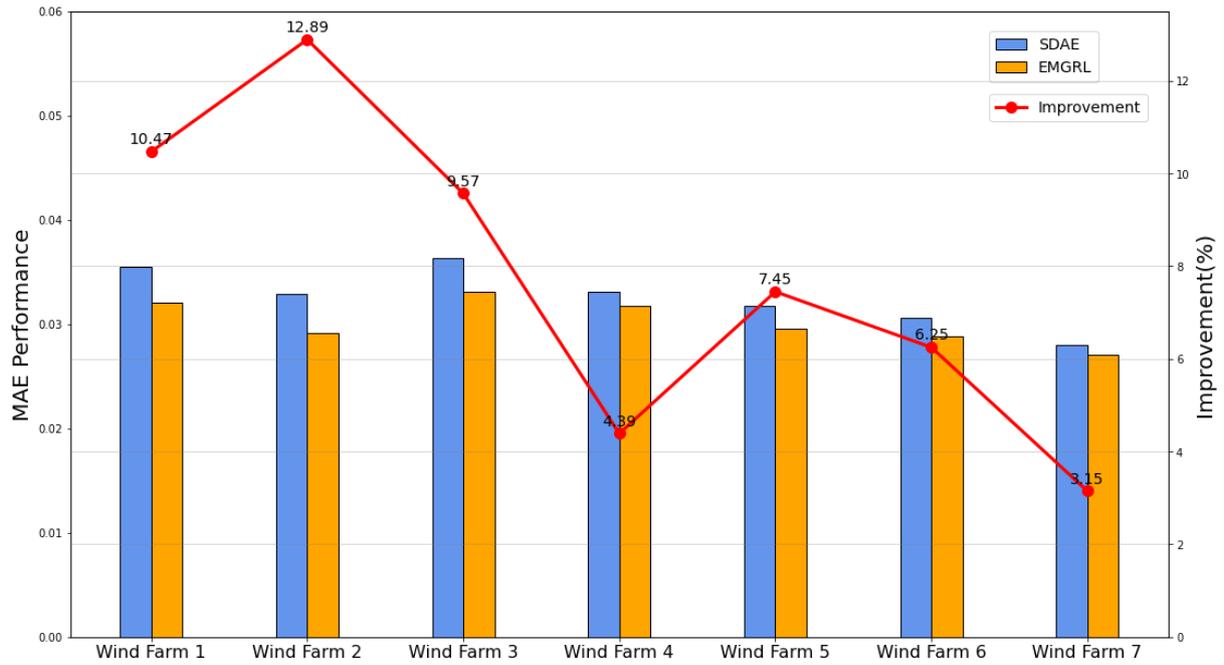

Figure 13: Improvement of EMGRL model on MAE compared to the best-performing benchmark model SDAE in GEFC Dataset.

Table 3: RMSE values of different baseline methods and EMGRL models on GEFC Datasets.

| Model | Wind Farm 1 | Wind Farm 2 | Wind Farm 3 | Wind Farm 4 | Wind Farm 5 | Wind Farm 6 | Wind Farm 7 |
|---|---|---|---|---|---|---|---|
| ARIMA | 0.2215 | 0.2087 | 0.2150 | 0.2073 | 0.1969 | 0.1986 | 0.1976 |
| SVR | 0.1887 | 0.1772 | 0.1891 | 0.1760 | 0.1712 | 0.1682 | 0.1593 |
| LightGBM | 0.1666 | 0.1702 | 0.1630 | 0.1661 | 0.1637 | 0.1550 | 0.1469 |
| LSTM | 0.1754 | 0.1668 | 0.1768 | 0.1637 | 0.1523 | 0.1545 | 0.1564 |
| SGNN | 0.1792 | 0.1622 | 0.1795 | 0.1761 | 0.1655 | 0.1617 | 0.1502 |
| SDAE | 0.1650 | 0.1582 | 0.1663 | 0.1496 | 0.1717 | 0.1501 | 0.1522 |
| COBRT | 0.1734 | 0.1821 | 0.2073 | 0.1784 | 0.2029 | 0.1685 | 0.1794 |
| EMGRL | **0.1505** | **0.1492** | **0.1605** | **0.1423** | **0.1628** | **0.1435** | **0.1473** |

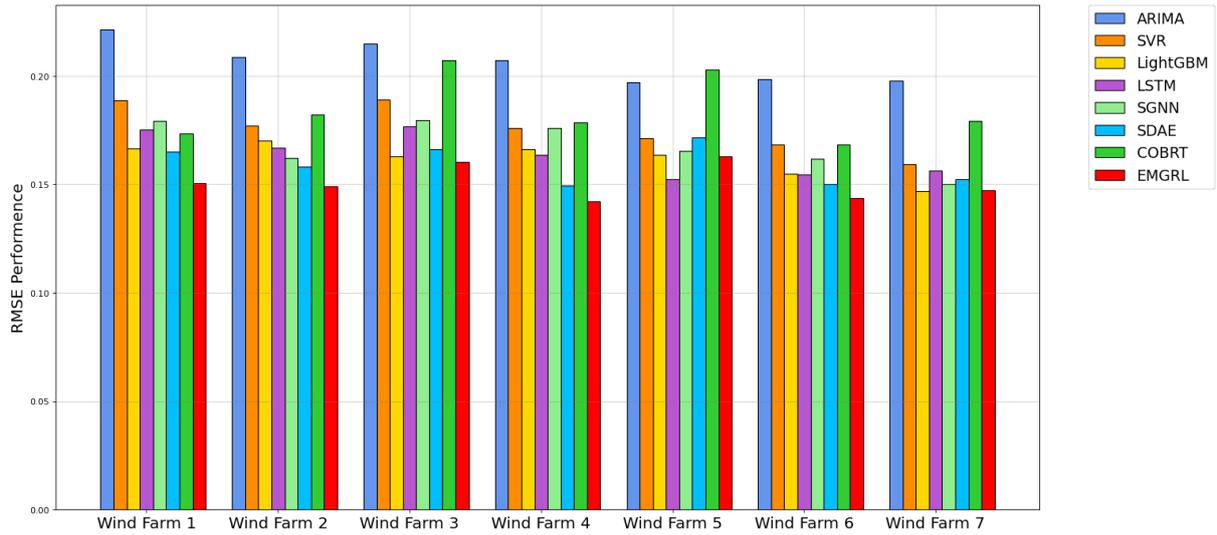

Figure 14: RMSE performance comparison between EMGRL and multiple baseline methods on GEFC Dataset.

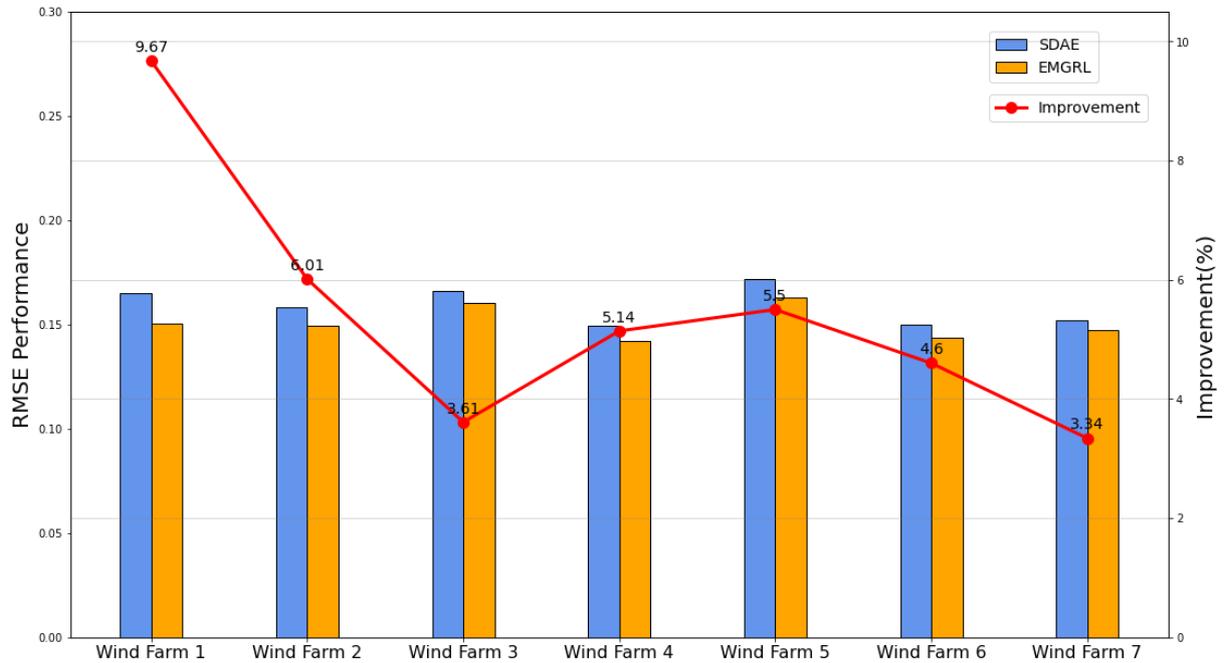

Figure 15: Improvement of EMGRL model on RMSE compared to the best-performing benchmark model SDAE in GEFC Dataset.

# 6 Conclusion

In this paper, we model all wind turbines within a wind farm as graph nodes in a wind farm graph constructed based on their geographical locations. We propose an ensemble model based on graph neural networks and reinforcement learning (EMGRL) for wind power forecasting (WPF). Our approach includes: (1) setting up a graph neural network to extract the spatial features of wind farm data; (2) generating an embedded vector that integrates model loss and Spatio-temporal data; and (3) obtaining effective wind power prediction results through ensemble learning. Comparative experiments on open datasets for WPF show that the EMGRL model outperforms the state-of-the-art (SOTA) baseline by 12.89%.

# References

Bludszuweit, H., Domı́nguez-Navarro, J. A., & Llombart, A. (2008). Statistical analysis of wind power forecast error. IEEE Transactions on Power Systems, 23, 983–991.

Bossanyi, E. (1985). Short-term wind prediction using kalman filters. Wind Engineering, 9, 1–8.

Chang, W.-Y. et al. (2014). A literature review of wind forecasting methods. Journal of Power and Energy Engineering, 2, 161.

Chen, C., & Liu, H. (2021). Dynamic ensemble wind speed prediction model based on hybrid deep reinforcement learning. Advanced Engineering Informatics, 48, 101290.

Council, G. W. E. (2022). Global wind report 2022. Global Wind Energy Council: Brussels, Belgium, .

Dayan, P., & Watkins, C. (1992). Q-learning. Machine learning, 8, 279–292.

Demolli, H., Dokuz, A. S., Ecemis, A., & Gokcek, M. (2019). Wind power forecasting based on daily wind speed data using machine learning algorithms. Energy Conversion and Management, 198, 111823.

Deng, X., Shao, H., Hu, C., Jiang, D., & Jiang, Y. (2020). Wind power forecasting methods based on deep learning: A survey. Computer Modeling in Engineering and Sciences, 122, 273.

Devi, A. S., Maragatham, G., Boopathi, K., & Rangaraj, A. (2020). Hourly day-ahead wind power forecasting with the eemd-cso-lstm-efg deep learning technique. Soft Computing, 24, 12391–12411.

Dhiman, H. S., Anand, P., & Deb, D. (2019). Wavelet transform and variants of svr with application in wind forecasting. In Innovations in Infrastructure (pp. 501–511). Springer.

Dong, X., Yu, Z., Cao, W., Shi, Y., & Ma, Q. (2020). A survey on ensemble learning. Frontiers of Computer Science, 14, 241–258.

Dong, Y., Zhang, H., Wang, C., & Zhou, X. (2021). A novel hybrid model based on bernstein polynomial with mixture of gaussians for wind power forecasting. Applied Energy, 286, 116545.

Draxl, C., Hodge, B., Clifton, A., & McCaa, J. (2015). Overview and meteorological validation of the wind integration national dataset toolkit. Technical Report National Renewable Energy Lab.(NREL), Golden, CO (United States).

Duan, J., Wang, P., Ma, W., Fang, S., & Hou, Z. (2022). A novel hybrid model based on nonlinear weighted combination for short-term wind power forecasting. International Journal of Electrical Power & Energy Systems, 134, 107452.

Du, P., Wang, J., Yang, W., & Niu, T. (2019). A novel hybrid model for short-term wind power forecasting. Applied Soft Computing, 80, 93–106.

Gupta, A., Sharma, K. C., Vijayvargia, A., & Bhakar, R. (2019). Very short term wind power prediction using hybrid univariate arima-garch model. In 2019 8th International Conference on Power Systems (ICPS) (pp. 1–6). IEEE.

Hanifi, S., Liu, X., Lin, Z., & Lotfian, S. (2020). A critical review of wind power forecasting methods—past, present and future. Energies, 13, 3764.

Hong, Y.-Y., & Rioflorido, C. L. P. P. (2019). A hybrid deep learning-based neural network for 24-h ahead wind power forecasting. Applied Energy, 250, 530–539.

Jalali, S. M. J., Osório, G. J., Ahmadian, S., Lotfi, M., Campos, V. M., Shafie-khah, M., Khosravi, A., & Catalão, J. P. (2021). New hybrid deep neural architectural search-based ensemble reinforcement learning strategy for wind power forecasting. IEEE Transactions on Industry Applications, 58, 15–27.

Jaseena, K., & Kovoor, B. C. (2021). Decomposition-based hybrid wind speed forecasting model using deep bidirectional lstm networks. Energy Conversion and Management, 234, 113944.

Jiading, J., Feng, W., Rui, T., Lingling, Z., & Xin, X. (2022). Ts xgb: Ultra-short-term wind power forecasting method based on fusion of time-spatial data and xgboost algorithm. Procedia Computer Science, 199, 1103–1111.

Jørgensen, K. L., & Shaker, H. R. (2020). Wind power forecasting using machine learning: State of the art, trends and challenges. In 2020 IEEE 8th International Conference on Smart Energy Grid Engineering (SEGE) (pp. 44–50). IEEE.

Kavasseri, R. G., & Seetharaman, K. (2009). Day-ahead wind speed forecasting using f-arima models. Renewable Energy, 34, 1388–1393.

Kisvari, A., Lin, Z., & Liu, X. (2021). Wind power forecasting–a data-driven method along with gated recurrent neural network. Renewable Energy, 163, 1895–1909.

Ko, M.-S., Lee, K., Kim, J.-K., Hong, C. W., Dong, Z. Y., & Hur, K. (2020). Deep concatenated residual network with bidirectional lstm for one-hour-ahead wind power forecasting. IEEE Transactions on Sustainable Energy, 12, 1321–1335.

Konda, V., & Tsitsiklis, J. (1999). Actor-critic algorithms. Advances in neural information processing systems, 12.

Lahouar, A., & Slama, J. B. H. (2017). Hour-ahead wind power forecast based on random forests. Renewable

energy, 109, 529–541.

Li, G., & Chiang, H.-D. (2016). Toward cost-oriented forecasting of wind power generation. IEEE Transactions on Smart Grid, 9, 2508–2517.

Lillicrap, T. P., Hunt, J. J., Pritzel, A., Heess, N., Erez, T., Tassa, Y., Silver, D., & Wierstra, D. (2015). Continuous control with deep reinforcement learning. arXiv preprint arXiv:1509.02971 , .

Liu, M., Cao, Z., Zhang, J., Wang, L., Huang, C., & Luo, X. (2020). Short-term wind speed forecasting based on the jaya-svm model. International Journal of Electrical Power & Energy Systems, 121, 106056.

Mahaseth, R., Kumar, N., Aggarwal, A., Tayal, A., Kumar, A., & Gupta, R. (2022). Short term wind power forecasting using k-nearest neighbour (knn). Journal of Information and Optimization Sciences, 43, 251–259.

Mnih, V., Kavukcuoglu, K., Silver, D., Rusu, A. A., Veness, J., Bellemare, M. G., Graves, A., Riedmiller, M., Fidjeland, A. K., Ostrovski, G. et al. (2015). Human-level control through deep reinforcement learning. nature, 518, 529–533.

Nelson, B. K. (1998). Time series analysis using autoregressive integrated moving average (arima) models. Academic emergency medicine, 5, 739–744.

Niu, Z., Yu, Z., Tang, W., Wu, Q., & Reformat, M. (2020). Wind power forecasting using attention-based gated recurrent unit network. Energy, 196, 117081.

Wang, G., Jia, R., Liu, J., & Zhang, H. (2020a). A hybrid wind power forecasting approach based on bayesian model averaging and ensemble learning. Renewable energy, 145, 2426–2434.

Wang, J., & Hu, J. (2015). A robust combination approach for short-term wind speed forecasting and analysis–combination of the arima (autoregressive integrated moving average), elm (extreme learning machine), svm (support vector machine) and lssvm (least square svm) forecasts using a gpr (gaussian process regression) model. Energy, 93, 41–56.

Wang, X., Guo, P., & Huang, X. (2011). A review of wind power forecasting models. Energy procedia, 12.

Wang, Y., Hu, Q., Meng, D., & Zhu, P. (2017). Deterministic and probabilistic wind power forecasting using a variational bayesian-based adaptive robust multi-kernel regression model. Applied energy, 208.

Wang, Y., Zou, R., Liu, F., Zhang, L., & Liu, Q. (2021). A review of wind speed and wind power forecasting with deep neural networks. Applied Energy, 304, 117766.

Welch, G. F. (2020). Kalman filter. Computer Vision: A Reference Guide, (pp. 1–3).

WHO. 2022. Available online: https://www.who.int/westernpacific/health-topics/air-pollution (accessed on 10 October 2022).

Wu, Q., Guan, F., Lv, C., & Huang, Y. (2021). Ultra-short-term multi-step wind power forecasting based on cnn-lstm. IET Renewable Power Generation, 15,.

Wu, Q., Zheng, H., Guo, X., & Liu, G. (2022). Promoting wind energy for sustainable development by precise wind speed prediction based on graph neural networks. Renewable Energy, .

Pelikan, E., Eben, K., Resler, J., Juruˇs, P., Krˇc, P., Brabec, M., Brabec, T., & Musilek, P. (2010). Wind power forecasting by an empirical model using nwp outputs. In 2010 9th international conference on environment and electrical engineering (pp. 45–48). IEEE.

Pole, A., West, M., & Harrison, J. (2018). Applied Bayesian forecasting and time series analysis. Chapman and Hall/CRC.

Ren, J., Yu, Z., Gao, G., Yu, G., & Yu, J. (2022). A cnn-lstm-lightgbm based short-term wind power prediction method based on attention mechanism. Energy Reports, 8, 437–443.

Salgado, P., Igrejas, G., & Afonso, P. (2018). Multi-kalman filter to wind power forecasting. In 2018 13th APCA International Conference on Automatic Control and Soft Computing (CONTROLO) (pp. 110–114). IEEE.

Santhosh, M., Venkaiah, C., & Vinod Kumar, D. (2020). Current advances and approaches in wind speed and wind power forecasting for improved renewable energy integration: A review. Engineering Reports, 2, e12178.

Shi, J., Qu, X., & Zeng, S. (2011). Short-term wind power generation forecasting: Direct versus indirect arima-based approaches. International Journal of Green Energy, 8, 100–112.

Shi, K., Qiao, Y., Zhao, W., Wang, Q., Liu, M., & Lu, Z. (2018). An improved random forest model of short-term wind-power forecasting to enhance accuracy, efficiency, and robustness. Wind energy, 21, 1383–1394.

Sideratos, G., & Hatziargyriou, N. D. (2007). An advanced statistical method for wind power forecasting. IEEE Transactions on power systems, 22, 258–265.

Sutton, R. S., Barto, A. G. et al. (1998). Introduction to reinforcement learning, .

Tian, S., Fu, Y., Ling, P., Wei, S., Liu, S., & Li, K. (2018). Wind power forecasting based on arima-lgarch model. In 2018 International Conference on Power System Technology (POWERCON) (pp. 1285–1289). IEEE.

Yan J., Zhang H., Liu Y., Han S., Li L., and Lu, Z.（2018）Forecasting the High Penetration of Wind Power on Multiple Scales Using Multi-to-Multi Mapping, IEEE Transactions on Power Systems, 33，3, 3276-3284.

Yatiyana, E., Rajakaruna, S., & Ghosh, A. (2017). Wind speed and direction forecasting for wind power generation using arima model. In 2017 Australasian Universities Power Engineering Conference (AUPEC) (pp. 1–6). IEEE.

Yu M., Zhang Z., Li X., Y J., G J., Liu Z., You B., Zheng X. and Yu R. (2020). Superposition Graph Neural Network for offshore wind power prediction. Future Generation Computer Systems, 113, 145-157.

Zeng, J., & Qiao, W. (2011). Support vector machine-based short-term wind power forecasting. In 2011 IEEE/PES power systems conference and exposition (pp. 1–8). IEEE.